# Ranking and Combining Latent Structured Predictive Scores without Labeled Data

Shiva Afshar[a], Yinghan Chen[b], Shizhong Han[c], Ying Lin[a]*

[a]*Department of Industrial Engineering, University of Houston, Houston, TX, 77204, USA;*

[b]*Department of Mathematics and Statistics, University of Nevada, Reno, NV, 89557, USA;*

[c]*Lieber Institute for Brain Development, Baltimore, MD, 21205, USA*

## Abstract

Combining multiple predictors obtained from distributed data sources to an accurate meta-learner is promising to achieve enhanced performance in lots of prediction problems. As the accuracy of each predictor is usually unknown, integrating the predictors to achieve better performance is challenging. Conventional ensemble learning methods assess the accuracy of predictors based on extensive labeled data. In practical applications, however, the acquisition of such labeled data can prove to be an arduous task. Furthermore, the predictors under consideration may exhibit high degrees of correlation, particularly when similar data sources or machine learning algorithms were employed during their model training. In response to these challenges, this paper introduces a novel structured unsupervised ensemble learning model (SUEL) to exploit the dependency between a set of predictors with continuous predictive scores, rank the predictors without labeled data and combine them to an ensembled score with weights. Two novel correlation-based decomposition algorithms are further proposed to estimate the SUEL model, constrained quadratic optimization (SUEL.CQO) and matrix-factorization-based (SUEL.MF) approaches. The efficacy of the proposed methods is rigorously assessed through both simulation studies and real-world application of risk genes discovery. The results compellingly demonstrate that the proposed methods can efficiently integrate the dependent predictors to an ensemble model without the need of ground truth data.

## 1. Introduction

Combining multiple predictors obtained from distributed data sources or various machine learning algorithms to an accurate meta-learner is promising to achieve enhanced and robust performance in a lots of classification problems (Erekat, Servis, Madathil, & Khasawneh, 2020; Song & Liu, 2018; Wang, Wang, & Tang, 2021; Wang, Wang, Tang, & Zhang, 2023). For instance, in genetics, researchers seek to understand disease-associated genes or variants by consolidating multiple association statistics originating from various genome-wide association studies conducted across diverse large-scale populations. The integration of these association statistics furnishes complementary evidence, thereby enhancing the discovery of risk (Brueggeman, Koomar, & Michaelson, 2020; Cao et al., 2022). Similarly, in the realm of anomaly detection, there is a need to combine predictions from distributed detectors or models, constructed from data captured by local sensors or diverse datasets. This enables the detection of more robust anomalies at a system-level (Araya, Grolinger, ElYamany, Capretz, & Bitsuamlak, 2017; Tama & Lim, 2021; Tsogbaatar et al., 2020; Vanerio & Casas, 2017; J. Zhang, Li, Nai, Gu, & Sallam, 2019). Furthermore, the ability to combine predictors obtained from distributed and local data holds significant promise in facilitating the development of federated or privacy-preserving machine learning models, all without the necessity of accessing or sharing local datasets (Chung & Al Kontar, 2024; Yue, Kontar, & Gómez, 2024). However, achieving the optimal combination of predictors necessitates prior knowledge regarding the reliability or accuracy of each model, which is often unavailable in practical applications. In genetics, for example, our current knowledge on the disease-gene associations is still very limited and the effectiveness of the existing genome-wide association studies remain unknown. Consequently, there is a pressing need to develop computational algorithms that can effectively rank predictors with unknown accuracy and achieve their optimal combination.

Ensemble learning algorithms are most widely used approaches to combine multiple predictors to a meta-learner, which consistently outperforms individual predictors. These ensemble approaches are broadly categorized into supervised and unsupervised methods. In the supervised ensemble learning realm, the techniques are primarily centered around the ranking and combination of different predictors using ground truth information. These supervised ensemble learning methods can be further divided into four principal categories: stacking, blending, bagging, and boosting. Stacking and blending techniques operate by aggregating predictions from multiple machine learning algorithms, all trained on the same dataset. Predictions generated by various algorithms are treated as new features and feed into a supervised learning model to predict the final ensembled scores (Kalia, 2018; Pavlyshenko, 2018; Wu, Zhang, Jiao, Guo, & Hamoud, 2021; Yao, Zhang, Luo, Liu, & Ren, 2022). Bagging-based ensemble models, such as random forest, enhance prediction performance by reducing the prediction variance using ensemble of machine learning models (such as unpruned decision trees) that are trained in different subsets of the data. Then, these models are combined using voting or averaging (Breiman, 1996; Caltanissetta, Bertoli, & Colosimo, 2023; Qi, 2012). Boosting-based methods, such as Adaboost and Gradient Boosting, were developed to mitigate the bias of prediction. They do so by weighted combining sequentially estimated weak learners into a meta-learner, assigning higher weights to the more accurate weak learners (Chen & Guestrin, 2016; Gaw, Yousefi, & Gahrooei, 2022; Iranzad et al., 2022; Marudi, Ben-Gal, & Singer, 2022; Schapire, 2013).

However, it's worth noting that estimating these supervised ensemble learning models typically require the access to the entire training dataset, which prohibit them to be used when the original training datasets are distributed and protected. Moreover, supervised ensemble learning methods heavily rely on a substantial amount of the labeled data to achieve robust prediction performance. In practice, however, real-world problems often suffer from a scarcity of labeled data, which can significantly undermine the effectiveness of supervised learning techniques. The lack of labeled genes is regarded as a critical challenge to build the effective risk gene discovery model for most complex diseases (Craig, 2008). The Autism Spectrum Disorders (ASD), for example, is a neurodevelopmental condition that has a strong genetic basis (Devlin & Scherer, 2012). While the role of genetic factors has been firmly established in ASD, the knowledge of individual gene's contribution to ASD development is still far from complete (Weiner et al., 2017). To address these issues, unsupervised ensemble learning approaches have been developed (Shaham et al., 2016; Valpola & Karhunen, 2002; Yang, 2016; J. Zhang et al., 2019). These approaches enable the combination of several predictors when the original training data and ground truth information are not readily available. A pivotal step in unsupervised ensemble learning methods involves estimating the reliability of each predictor without labeled data. Various approaches for estimating predictor reliability without labeled data have been proposed in the literature. For example, building upon the model presented in (Dawid & Skene, 1979; Jaffe, Nadler, & Kluger, 2015; Y. Zhang, Chen, Zhou, & Jordan, 2014), recent works have employed methods such as maximum likelihood estimation and the EM algorithm. These techniques utilize an iterative approach to solve the maximum likelihood estimation of parameters related to the latent probability distribution of predictors (Parisi, Strino, Nadler, & Kluger, 2014). In (Traganitis, Pages-Zamora, & Giannakis, 2018), a maximum posterior approach was introduced to estimate the confusion matrix of predictors in a multi-class ensemble classification problem. Additionally, (Li & Yu, 2014) and (Cabrera-Bean, Díaz-Vilor, & Vidal, 2016) derived an upper bound for the error rate in EM-based and maximum posterior-based approaches for determining the unknown accuracy of each predictor. However, it's important to note that the EM-based approach, while conceptually straightforward, has limitations. It only guarantees convergence to a local optimal solution and can entail high computational costs. To mitigate this issue, a spectral-based approach was recently developed by spectrally decomposing the covariance matrix of predictors and inferring the model parameters from the covariance matrix (Ionita-Laza, McCallum, Xu, & Buxbaum, 2016; Jain & Oh, 2014). Nonetheless, one limitation of spectral-based approaches is their assumption of conditional independence among classifiers given the true label. This assumption may not hold in many real-world problems. Recent research has aimed to relax this conditional independence assumption by considering a latent tree structure capable of capturing dependencies between predictors (Jaffe, Fetaya, Nadler, Jiang, & Kluger, 2016; Kleindessner & Awasthi, 2018; Shaham et al., 2016). However, it's essential to acknowledge that these studies primarily focused on integrating the categorical outcomes predicted from the models, which cannot be used to deal with the continuous predictive scores often encountered in the real-world problems. The advantages of using continuous predictive scores compared with categorical outcomes include the better characterization of prediction uncertainty and the independence of cut-off threshold.

Hence, there arises a critical necessity to develop new models capable of ranking and combining the dependent and continuous predictive scores without labeled data. To fill this gap, this paper introduces an innovative Structured Unsupervised Ensemble Learning model (SUEL) accompanied by two efficient learning algorithms to estimate the model (denoted as SUEL.CQO and SUEL.MF). The proposed SUEL model 1) provides an optimal ranking and combination schema of continuous predictive scores without labeled data; 2) explicitly characterizes the dependencies between predictive scores through a set of latent models; 3) efficiently uncovers the latent models and estimates model parameters by leveraging the spectral structure of the correlation coefficient matrix between predictors. The novelty and contributions of the proposed SUEL methods can be succinctly summarized as follows:

- A novel structured unsupervised ensemble learning model is proposed to rank and combine the dependent and continuous predictive scores without labeled data.
- Two novel learning algorithms based on the spectral structure of correlation coefficient matrix between predictive scores are developed to efficiently estimated the proposed model.
- Properties of the proposed model and algorithms are analysed theoretically, including how to determine the optimal latent structure among dependent predictive scores and the optimal weights for combining the scores.
- The proposed method is applied to a risk gene discovery for autism spectrum disorder problem, in which the proposed method optimally combines the existing disease-association evidence to an unsupervised ensemble score with enhanced prediction accuracy.

The rest of the paper is organized as follows. Section 2 introduces the proposed SUEL model, elaborates its theoretical properties and presents two correlation matrix-based decomposition algorithms to solve it. Section 3 conducts a simulation study to evaluate the performance of the proposed method. A real-world case study on risk genes discovery is further presented in Section 4. The conclusions are drawn in Section 5.

## 2. Materials and methods

*2.1 Problem setup*

We consider the problem of ranking and combining $M$ predictors with unknown accuracy, denoted as $\{f_i\}_{i=1}^M$, to predict a binary outcome. Each predictor provides continuous predictive scores $f_i(x_t)$ for a set of $N$ samples, denoted as $D = \{x_t\}_{t=1}^N$. It is notable that the feature information used in each predictor of $N$ samples can be different. The true outcomes of these samples are unknown and represented as a binary variable, i.e., $\{y_t\}_{t=1}^N$. Each predictor is trained on its local training dataset which is also unavailable to us. To rank and combine the predictors to a meta-learner, we adopt the ensemble learning approach which weighted aggregates the predictors to an ensemble model, denoted as $g(x_t) = \sum_{i=1}^M w_i f_i(x_t), t \in \{1, \ldots, N\}$, where $w_i$ represents the ensemble weight assigned to $i$th predictor and a higher weight indicates the predictor is more reliable or accurate. The objective is to find a set of weights that optimally rank and combine the predictors to best predict the distribution of outcome variable without labeled data.

As the scores predicted from $M$ predictors are continuous, a set of two-component Gaussian mixture distributions are used to represent the relationships between the predictors and outcome variable (Ionita-Laza et al., 2016). If the scores do not follow Gaussian distribution assumption, the Cox-Box transformation (Daimon, 2011) can be used to transform the scores to Gaussian distributions first and the proposed method can be applied on the transformed scores. The assumption is that the scores predicted from each predictor have a two-component Gaussian mixture distribution where each component represents the distribution of the predicted scores in one class. The Gaussian mixture distribution can be formulated as:

$$P(f_i(x_t)) = \pi P(f_i(x_t)|y_t = 1) + (1 - \pi)P(f_i(x_t)|y_t = 0) \tag{1}$$

where $\pi$ represents the prior probability of belonging to class 1. $P(f_i(x_t)|y_t = 1)$ and $P(f_i(x_t)|y_t = 0)$ represent the component-specific distributions of the predicted score in two classes, respectively. The Gaussian distributions in two classes are described by component-specific means, $\mu_{i1}$ and $\mu_{i0}$, and component-specific standard deviations, $\sigma_{i1}$ and $\sigma_{i0}$.

Conventional ranking and combining methods assume the prediction scores are conditionally independent given the true label of outcome and the variances of two components are the same, i.e., $\sigma_{i1} = \sigma_{i0} = \sigma_i$. Based on these assumptions, the maximum likelihood estimation of the ensemble weights can be explicitly characterized in **Property 1**.

***Property 1*** *– If the predictive scores from each predictor follow a two-component Gaussian mixture distribution with different component-specific means but equal variance, the maximum likelihood estimation of the ensemble weights equals to:*

$$w_i = \frac{1}{2\sigma_i^2}(\mu_{i1} - \mu_{i0}). \tag{2}$$

**Property 1** highlights that the scores exhibiting a larger discrepancy between component-specific means and lower variance are indicative of greater reliability and accuracy. The larger difference between component-specific means indicates a better discrimination between the two classes, while lower variance suggests that the predicted scores within each class are more closely clustered. Figure 1 illustrates this concept with an example, showcasing how the difference between component-specific means and the variance of a Gaussian mixture distribution can impact predictor performance. As depicted, a larger difference in component-specific means and smaller component-specific standard deviation correspond to enhanced prediction performance of the score. It's worth noting that the maximum likelihood estimation of ensemble weights proposed in **Property 1** holds an advantage over weights defined in conventional ranking and combining methods (Ionita-Laza et al., 2016). These conventional methods solely consider the difference between component-specific means and disregard the influence of component-specific variance on predictor performance and, subsequently, the estimation of ensemble weights.

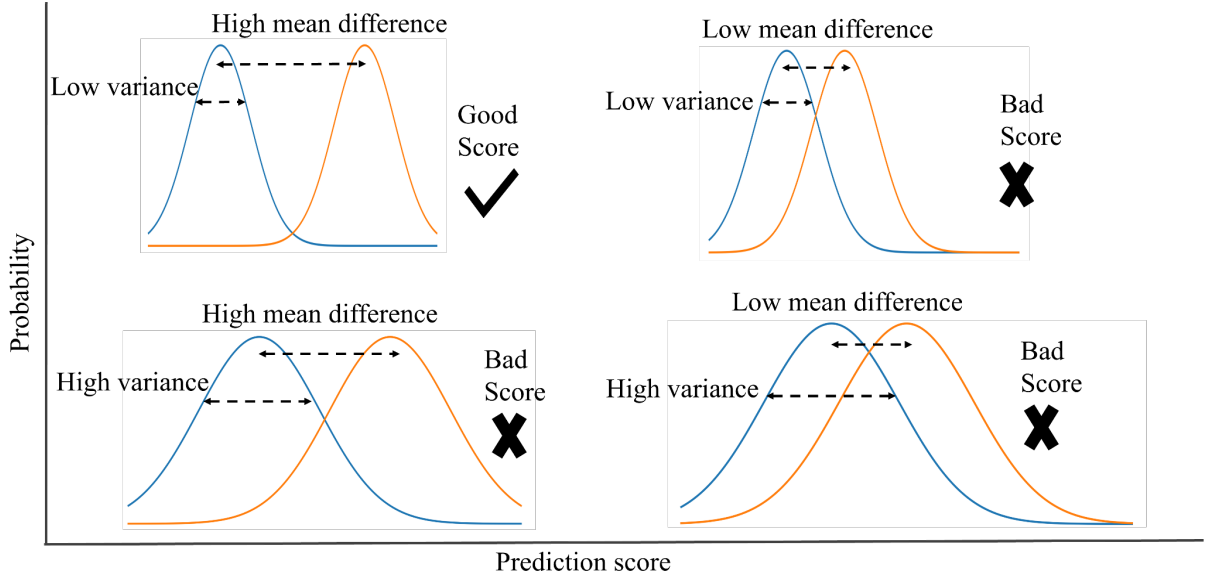

Figure 1. The impact of means and variance of components in accuracy of predictors.

## *2.2 The proposed structured unsupervised ensemble learning method for dependent scores*

Conventional ranking and combining methods often hinge on the assumption that predictors are independent, given the outcome variable. This assumption is known as the conditional independence assumption. However, in practical scenarios, predictors can exhibit dependencies. This dependence can arise when predictors are estimated using overlapping training data or similar algorithms, challenging the conditional independence assumption. To capture the dependency among $M$ predictors, a novel structured unsupervised ensemble learning model is proposed by assuming the predictors are correlated in $K$ ($K \leq M$) latent models (as shown in Figure 2). The predictors generated from the same latent model tend to exhibit high correlations, while those stemming from different latent models are conditionally independent given the true label of outcome. Conceptually, each latent model can be viewed as a distinct source of information or an algorithm, and the associated predictors are essentially different perturbations or variations of this model. The latent models are conditionally independent of each other when considering the true label of the outcome. Each latent model, denoted as $h_k$, $k = 1, \ldots, K$, is described by an underlining two-component Gaussian mixture distribution. Thus, the distribution of each latent model can be represented as:

$$h_k(\cdot \,|y_t = 0) \sim N(\mu_k^0, \sigma_k^2), \quad h_k(\cdot \,|y_t = 1) \sim N(\mu_k^1, \sigma_k^2) \qquad (3)$$

where $\mu_k^0$ and $\mu_k^1$ represent the component-specific means for the latent model distribution and $\sigma_k^2$ represents the component-consistent variance of the distribution.

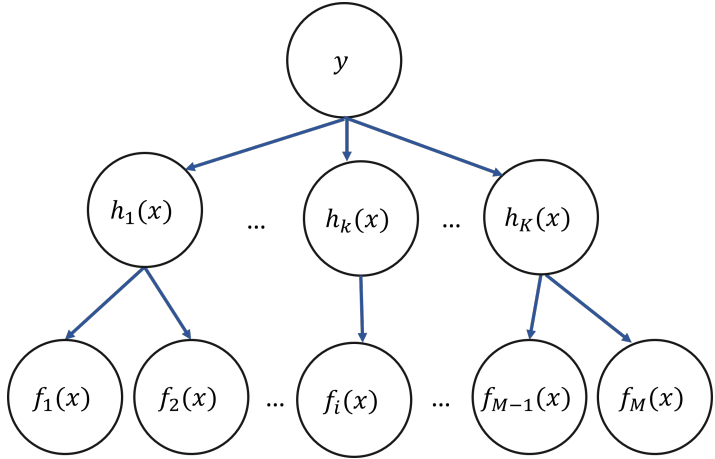

Figure 2. The structure of the proposed method.

The set of predictors that are associated with a latent model, $h_k$, is represented as $\{f_i\}_{i \in C_k}$, where $C_k$ indicates the set of associated predictors. Each predictor is assumed to be a perturbation of the latent variable following the linear relationship below:

$$f_i(x_t) = b_i h_k(x_t) + \varepsilon_i, \; \varepsilon_i \sim N(0, \theta_i^2) \tag{4}$$

where $b_i$ represents the strength of association between the predictor and corresponding latent model and $\varepsilon_i$ represents a normally distributed random noise that is independent of the latent model. $\theta_i^2$ indicates the variance of the random noise for $i$th score.

## 2.3 Property of the proposed model

This session discusses the statistical properties of the proposed structured unsupervised ensemble learning (SUEL) model. These statistical properties encompass: 1) the optimal weights of standardized predictive scores under the latent group structure, which is provided in **Property 2**; 2) the covariance matrix of the standardized predictors, if arranged under certain order, has a block-wise structure with each block is a rank-one submatrix (**Lemma 1**); 3) the number of the latent models in the proposed SUEL model can be determined by the eigen decomposition of the covariance matrix between standardized predictors (**Lemma 2**).

***Property 2*** – *Given M predictors that are correlated by the latent group structure described in the proposed SUEL model and have standardized predictions, the weight of ith predictor ($w_i$) can be estimated as:* $w_i = -\frac{b_i \mu_k^0}{\pi - (1-\pi) b_i^2 {\mu_k^0}^2}$.

Proof of **Property 2** is provided in the Supplemental materials.

It is notable that the numerator in the equation of **Property 2** represents the difference in component-specific means of $f_i(x)$ and the denominator represents the total variance of $f_i(x)$. Compared to the accuracy of conditionally independent predictors obtained from the literature ($w_i = -\frac{\mu_{i0}}{\pi}$) (Ionita-Laza et al., 2016), the accuracy of predictors in the proposed method depends on 1) the accuracy or reliability of its associated latent model ($\mu_k^0$), 2) the strength of association between that predictor with the latent model ($b_i$) and 3) the variance of the predictor ($b_i^2 \sigma_k^2 + \theta_i^2$). The variance of each predictor can also be estimated using the parameters $\mu_k^0$ and $b_i$. Therefore, we need to estimate $\mu_k^0$, $b_i$, and $\pi$ to compute $w_i$ for each predictor.

If the latent group structure is known and $C_k$ denotes the set of predictors that are associated to latent model $k$, the covariance matrix of $M$ standardized predictors has block-wise structure as illustrated in **Lemma 1**.

***Lemma 1*** – *Given the covariance matrix among M standardized predictors as* $\{r_{ij} = E[(f_i(x) - E[f_i(x)])(f_j(x) - E[f_j(x)])]\}_{i,j=1,..,M}$, *the off-diagonal elements of covariance matrix has a block-wise structure with each block is a rank-one submatrix and can be represented by the following piecewise function:*

$$r_{ij} = \begin{cases} v_i^{on} v_j^{on} & \text{if } i \neq j \text{ and}, f_i \text{ and } f_j \text{ are associated with the same latent model} \\ v_i^{off} v_j^{off} & \text{if } f_i \text{ and } f_j \text{ are associated with different latent models} \end{cases}$$

where $v^{on} = [\sigma_k b_i]_{k \in \{1,...,K\}, i \in C_k}$, $v^{off} = \left[\sqrt{\frac{1-\pi}{\pi}} \mu_k b_i\right]_{k \in \{1,...,K\}, i \in C_k}$.

Proof of **Lemma 1** is provided in the Supplemental materials.

It can be observed from **Lemma 1** that the covariance matrix is a combination of 2 rank-one matrices. The elements in diagonal blocks, i.e. correlations between predictors associated with the same latent model, correspond to the rank-one matrix $v^{on}(v^{on})^T$, which only depends on the strength of association between the predictors and their corresponding latent models, $\boldsymbol{b}_k = \{b_i\}_{i \in C_k}$, and the standard deviation of latent model ($\sigma_k$). The elements in the off-diagonal blocks, which represent the correlations between the conditional independent predictors, correspond to another rank-one matrix $v^{off}(v^{off})^T$. It depends on the strength of association between the predictors and their corresponding latent models and the difference in the component-specific means of the associated latent model ($\mu_k$). Using a binary variable to indicate whether a pair of predictors are associated with the $k$th latent model or not, denoted as $I_k(i,j)$, the covariance matrix in **Lemma 1** can be rewritten as Equation 5 below:

$$r_{ij} = I_k(i,j)v^{on}(v^{on})^T + (1 - I_k(i,j))v^{off}(v^{off})^T, \quad \forall i \neq j. \tag{5}$$

**Lemma 1** also provides insight into the fact that if we have knowledge of the assignment of scores to the latent models and the optimal number of latent models, we can decompose the covariance matrix. This decomposition allows us to estimate the parameters in the proposed method based on the elements within the covariance matrix. Determining the optimal number of latent models is subsequently addressed in **Lemma 2**.

***Lemma 2** – The covariance matrix of M predictors ( R ) can be decomposed as $R = G + \Theta$. G is a rank-K matrix relating to the covariance of latent models and $\Theta$ is a diagonal matrix represents the variances of random noise. If the variance of latent models is larger than the variance of random noise, the number of the latent models (K) can be determined by the number of leading components in the eigen decomposition of the covariance matrix (R).*

Proof of **Lemma 2** is provided in the Supplemental materials.

According to **Lemma 2**, to determine the number of the latent models in a set of standardized predictors, the sample covariance matrix needs to be first calculated. Then, the number of leading eigenvalues can be estimated by using the principal component analysis on the sample covariance matrix (Mishra et al., 2017).

*2.4 Solving the structured unsupervised ensemble learning model via two novels correlation-based algorithms*

The proposed structured unsupervised ensemble learning method can be characterized as a statistical model involving latent variables. While the Expectation-Maximization (EM) algorithm is a conventional method for estimating models with latent variables, it has its limitations. EM's performance is notably sensitive to the initial values, and it can be computationally expensive, particularly when dealing with a large number of parameters. To address these challenges and arrive at a unique optimal solution for the proposed model while maintaining computational efficiency, we introduce two innovative correlation-based decomposition algorithms.

The proposed algorithms consist of two sequential steps: optimal assignment of predictors to latent groups and estimation of unknown parameters in the model. The initial steps of both algorithms are identical. They involve standardizing the scores obtained from predictors and determining the optimal latent structure of the predictors. This includes identifying the number of latent models and assigning predictors to these latent models. This is achieved through the eigen decomposition of the covariance matrix, as elucidated in **Lemma 2**, and the use of a quartet-based similarity matrix, as explained in Section 2.5. The quartet-based similarity matrix is derived from the covariance matrix and is demonstrated to be effective in accurately distinguishing between predictors associated with the same latent model and those associated with different latent models, as highlighted in **Lemma 3**.

Using the optimal latent structure identified through the quartet-based similarity matrix, we can infer the parameters in the proposed method. These inferences rely on the relationships between model parameters and elements in the covariance matrix, as illustrated in **Lemma 1**. However, estimating these parameters directly from the covariance matrix poses challenges, as it constitutes a nonconvex problem for which no closed-form solution exists.

To tackle this issue, we introduce two algorithms in the second step. The first algorithm, denoted as SUEL.CQO, approximates the covariance matrix using a set of exponential functions and subsequently estimates the parameters through the solution of a constrained quadratic optimization problem. This process is detailed in Section 2.6.

Recognizing that the exponential approximation of the covariance matrix may introduce significant bias, we have introduced an alternative algorithm, SUEL.MF, detailed in Section 2.7. This matrix-factorization-based approach is designed to achieve more accurate estimation of the model's parameters. It decomposes the covariance matrix, as demonstrated in **Lemma 2**, and estimates the model parameters using a coordinate descent algorithm. Ultimately, the weight assigned to each predictor is estimated based on **Property 2**. These weights are then employed to construct the ensemble model.

*2.5 Determining the optimal latent structure using quartet-based similarity matrix*

The first step in our proposed method involves determining the optimal number of latent models and assigning predictive scores to each latent model. To achieve this, we introduce a quartet-based similarity matrix, which facilitates the identification of the global optimal assignments of predictive scores to latent models with relatively low computational cost (as outlined in Algorithm 1).

The quartet-based similarity matrix is constructed based on correlations between scores. Our theoretical analysis demonstrates that dependent predictors, those associated with the same latent model, tend to exhibit larger similarities in the quartet-based similarity matrix compared to conditionally independent predictors, particularly when the accuracy of predictors is sufficiently high. Consequently, we can derive the optimal assignments of predictors by employing hierarchical clustering on the quartet-based similarity matrix, grouping similar objects into the same cluster. Specifically, the quartets of a covariance matrix are represented as the determinants of every 2×2 submatrix.

***Definition 1*** – *The quartet of four predictors, $f_i(x)$, $f_j(x)$, $f_k(x)$, $f_l(x)$, is defined as* $U_{ijkl} = \det\begin{pmatrix} r_{ij} & r_{il} \\ r_{kj} & r_{kl} \end{pmatrix}$.

Based on the low rank structure of covariance matrix described in **Lemma 1**, the quartets defined above are sparse, with the following property.

***Property 3*** – $U_{ijkl} = 0$ *if and only if three or more of the predictors ($i, j, k,$ and $l$) are associated with the same latent model or they belong to four different latent models.*

Proof of **Property 3** is provided in the Supplemental materials.

To detect the predictors strongly violating the conditional independence assumption and fall into the same latent model, the quartet-based similarity between each pair of predictors, denoted as $s_{ij}$, is defined by aggregating all quartets including this pair of predictors.

**Definition 2** – *The quartet-based similarity between predictors $f_i(x)$ and $f_j(x)$ is defined as:* $s_{ij} = \sum_{k \neq i,j; l \neq i,j} U_{ijkl}$.

Intuitively, each element of the quartet-based similarity matrix represents the similarity between a pair of predictors. The large similarity indicates that the two predictors are more likely to be correlated by the same latent model. When $f_i(x)$ and $f_j(x)$ are associated with different latent models, the similarity between them is estimated by aggregating their associated quartets, most of which come from four different latent models and thus have values equal to zero (**Property 3**). This fact leads to the small value of similarity.

When $f_i(x)$ and $f_j(x)$ are associated with the same latent model, on the other hand, their similarity is large because most of the associated quartets have non-zero and positive values. Therefore, the similarity between predictors associated with the same latent model is always higher than the similarity of conditional independent predictors if the accuracy of predictors is high enough, which is rigorously proved in **Lemma 3**.

Assuming 1) the prior distribution of two classes is balanced, i.e. $\pi = 0.5$; 2) the absolute value of conditional covariance between dependent predictors are lower bounded by $\theta$, i.e. $|cov(f_i(x), f_j(x)|y = 1)| \geq \theta$ and $|cov(f_i(x), f_j(x)|y = 0)| \geq \theta$, and 3) the accuracy of predictors are lowered bounded by $\delta$, i.e. $|w_j| \geq \delta$, the following lemma can be obtained:

**Lemma 3-** *Under the assumption described above if $f_i(x)$ and $f_j(x)$ are associated with the same latent model then*

$$s_{ij} > \frac{1}{4}\theta\delta^2 M^2 \left(1 - \frac{3}{K} + \frac{2}{K^2}\right)$$

*and if $f_i(x)$ and $f_j(x)$ are associated with different latent models, then*

$$s_{ij} < \frac{M^2}{K}\left(5 - \frac{6}{K}\right).$$

Proof of **Lemma 3** is provided in the Supplemental materials.

From **Lemma 3,** we can draw the conclusion that if $\delta^2 \geq \frac{20}{\theta(K-2)}$, where $\delta^2$ represents a threshold of predictor accuracy, the lower bound of the quartet-based similarity between two predictors associated with the same latent model exceeds the upper bound of the quartet-based similarity between two predictors associated with different latent models. Consequently, the similarities between conditionally independent predictors are consistently lower than the similarities between conditionally dependent predictors. The values in similarity matrix exhibit a clear gap that even a simple single-linkage hierarchical clustering algorithm can recover the correct group structure of predictors. Therefore, the hierarchical clustering algorithm can be used. Utilizing hierarchical clustering on the predictors, we can identify groups of predictors that exhibit similar quartet-based similarity values, signifying that they belong to the same latent model. If the hierarchical clustering algorithm does not work, more robust clustering algorithms, such as the spectral clustering, can be used. Thus, the process of determining the number of latent models and assigning predictors to each latent model is succinctly outlined in Algorithm 1.

**Algorithm 1: Estimating the number of latent models (K) and assignments of scores to latent models ($C_k$s)**

**Input:** Prediction scores from $M$ predictors on $N$ samples ($f_i(x_t), i \in \{1, ..., M\}, t \in \{1, ..., N\}$)
1. Standardize the prediction scores.
2. Estimate the covariance matrix between standardized scores.
3. Estimate the number of latent groups by computing the rank of estimated covariance matrix using the optimal number of components in principal component analysis (PCA).
4. Obtain the quartet-based similarity matrix ($\hat{S}$).
5. Estimate $C_k$s by performing hierarchical clustering on the quartet-based similarity matrix.

**Output:** Number of latent groups ($K$) and the assignments ($C_k, k \in \{1, ..., K\}$)

## 2.6 Estimating the model parameters using a constrained quadratic optimization (SUEL.CQO) problem

We are interested in estimating the model parameters, including the component-specific means and variance in the latent models ($\mu_k^0, k = 1, ..., K$), strength of correlation between the predictors and the latent models ($b_i, i = 1, ..., M$), and the prior distribution of outcome variable ($\pi$), to infer the accuracy of each score. Given that the covariance matrix between standardized predictors can be expressed as a piecewise function concerning the model parameters, we employ a collection of linear equations to approximate the logarithm of the covariance matrix. Subsequently, we address the estimation of these unknown parameters by resolving a constrained quadratic optimization problem.

Specifically, the covariance matrix has a block-wise structure where the diagonal blocks represent the covariance among the scores that are grouped to the same latent model and off-diagonal blocks represent the covariance between the scores that are conditionally independent. According to **Lemma 1**, elements in the covariance matrix can be represented by a piecewise function on the model parameters. Based on these relationships, the elements in the diagonal blocks of the covariance matrix are decomposed as $|r_{ij}| = e^{s_o}.e^{d_i}.e^{d_j}$, with $e^{s_o}, e^{d_i}$, and $e^{d_j}$ represent $|b_i|, |b_j|$, and $\sigma_o^2$ respectively, $\forall i, j \in C_k, k = 1, ..., K$. In the off-diagonal blocks of the covariance matrix (i.e. $i \in C_v, j \in C_u, v \neq u, u, v \in \{1, ..., K\}$), the elements are decomposed as $|r_{ij}| = e^p.e^{d_i}.e^{d_j}.e^{l_v}.e^{l_u}$, where $e^p, e^{d_i}, e^{d_j}$, $e^{l_v}$, and $e^{l_u}$ present $\frac{1-\pi}{\pi}, |b_i|, |b_j|, |\mu_v^0|$, and $|\mu_u^0|$, respectively. Each element in the covariance matrix provides a relationship or equation between the model parameters. Therefore, we obtain a system of $\frac{M(M-1)}{2}$ linear equations that capture the relationships between $M + 2K + 1$ model parameters. Since the population covariance matrix ($R$) is unknown, sample covariance matrix ($\hat{R}$) is used as an unbiased estimation. Thus, the relationships between model parameters and elements in the covariance matrix is formulated as a multivariate linear equation problem with random noises in the following:

$$\begin{cases} \log|\hat{r}_{ij}| = s_o + d_i + d_j + \epsilon & i,j \in C_t \\ \log|\hat{r}_{ij}| = p + d_i + d_j + l_v + l_u + \epsilon & i \in C_v, j \in C_u, \end{cases} \quad (6)$$

where $\epsilon$ represents the random noise in sample covariance matrix. The problem can be reformulated as $\boldsymbol{z} = A\boldsymbol{g} + \boldsymbol{\epsilon}$, where $\boldsymbol{z}$ is a $\frac{M(M-1)}{2}$ dimensional vector representing log of the upper triangular part of the sample covariance matrix, $A$ is a $\frac{M(M-1)}{2} \times (M + 2K + 1)$ index matrix provides the indexes of each parameter in each equation. $\boldsymbol{g}$ is a $M + 2K + 1$ dimensional vector including all unknown parameters. As the value of $\log|\widehat{r_{ij}}|$ is always nonpositive, we require the estimated parameters to be less than or equal to 0. The constrained quadratic optimization problem is formulated as:

$$\underset{\boldsymbol{g}}{\text{minimize}} \|A\boldsymbol{g} - \boldsymbol{z}\|_2^2, \text{ subject to } \boldsymbol{g} \leq \boldsymbol{0}. \quad (7)$$

The solutions of Equation 5.7 can be efficiently obtained and transferred back to the model parameters using $\pi = \frac{1}{1+e^p}$, $|b_i| = e^{d_i}$, and $|\mu_v^0| = e^{l_v}$. The sign of the estimated accuracy for each score is determined by the sign of its correlation coefficient with other scores. Algorithm 2 summarizes the steps of SUEL.CQO.

---

**Algorithm 2: Estimating the model parameters required to compute ensemble weights using SUEL.CQO**

---

**Input:** The assignments ($C_k, k \in \{1, ..., K\}$), sample covariance matrix ($\hat{R}$), prediction scores ($f_i(x_t), i \in \{1, ..., M\}, t \in \{1, ..., N\}$), number of latent groups ($K$)

1. Formulate the linear system of equations as Equation 5.6.
2. Solve the constrained quadratic optimization problem in Equation 5.7 to estimate the model parameters ($\hat{\pi}, \hat{b}, \widehat{\mu_0}$).
3. Estimate the accuracy of each prediction score as $w_i = -\frac{b_i \mu_k^0}{\pi - (1-\pi) b_i^2 {\mu_k^0}^2}$.

**Output:** Compute the ensemble scores as $g(x_t) = \sum_{i=1}^{M} w_i f_i(x_t)$.

---

## *2.7 Estimating the model parameters using a matrix-factorization-based (SUEL.MF) approach*

While the SUEL.CQO algorithm offers a more cost-effective means of estimating model parameters in comparison to the EM algorithm, there are potential issues with the accuracy of the parameters estimated through this approach. This can be attributed to:1) the logarithm transformation of covariance matrix reduces the magnitude of estimation bias; 2) The presence of bias in the constrained quadratic optimization problem, particularly when the number of equations in the linear system of equations exceeds the number of variables. To address these issues, we further proposed a matrix-factorization-based algorithm (SUEL.MF). This algorithm directly decomposes the covariance matrix to estimate unknown parameters and has the potential to provide more accurate estimations compared to SUEL.CQO.

According to **Lemma 2**, the covariance matrix can be decomposed to three matrices, the coefficients between scores and latent models ($B$), The covariance of latent models ($\Lambda$), and a diagonal matrix that represents the variance of random noise ($\Theta$), as shown in Equation 8. Therefore, by estimating $B$, $\Lambda$, and $\Theta$, we can estimate the parameters that are required to compute the reliability of each predictor ($w_i$). Representing the coefficients between scores and latent models as a sparse coefficient matrix $B$, i.e. $B = \begin{bmatrix} \boldsymbol{b}_1 & \cdots & 0 \\ \vdots & \ddots & \vdots \\ 0 & \cdots & \boldsymbol{b}_K \end{bmatrix}_{M \times K}$ with $\boldsymbol{b}_i$ indicates the vector of coefficients associated with $i$th latent model, the covariance of latent models as $\Lambda = \begin{bmatrix} \sigma_1^2 & \cdots & \frac{1-\pi}{\pi} \mu_1^0 \mu_K^0 \\ \vdots & \ddots & \vdots \\ \frac{1-\pi}{\pi} \mu_1^0 \mu_K^0 & \cdots & \sigma_K^2 \end{bmatrix}_{K \times K}$ and the variance of random noise as $\Theta = \begin{bmatrix} \theta_1^2 & \cdots & 0 \\ \vdots & \ddots & \vdots \\ 0 & \cdots & \theta_M^2 \end{bmatrix}_{M \times M}$, the covariance matrix between predictors can be decomposed as:

$$R = B\Lambda B^T + \Theta \qquad (8)$$

The $\Lambda$ matrix can be further decomposed to eigenvectors $U$ and eigenvalues $V$, where the eigenvectors are related to the unknown parameters $\mu_1^0, \ldots, \mu_K^0$. To estimate the unknown parameters $B$, $U$, $V$, and $\Theta$, the following optimization problem needs to be solved:

$$\text{minimize } \|R - BUV(BU)^T + \Theta\|_F^2,$$

$$\text{subject to } UU^T = I \text{ and } B \text{ is sparse with off-diagonal blocks are 0.} \qquad (9)$$

To solve the optimization problem proposed formulated in Equation 9, a block coordinate descent (BCD)-based algorithm is introduced. The BCD algorithms break down a given matrix-based nonconvex optimization problem into convex subproblems (based on blocks of columns) which are easier to solve. To solve the problem presented in Equation 9, we adopt an iterative two-stage approach that estimate $\Theta$ and then estimate $B$ and $U$ alternatively.

**Estimation of $\Theta$:** When $B$ and $U$ are fixed, the diagonal matrix $\Theta$ can be estimated via $\Theta = diag(R - BUV(BU)^T)$.

**Estimation of $B$ and $U$:** When $\Theta$ is fixed, the estimation of $B$ and $U$ can be obtained by the following steps. First, we decompose the partial residual matrix, $R - \Theta$, to its eigenvectors ($W$) and eigenvalues ($V$), where the columns of $W$ contain the first $K$ leading eigenvectors of the partial residual matrix. Then, the parameters $B$ and $U$ can be estimated by further decomposing the matrix of first $K$ leading eigenvectors to a sparse and structured matrix and an orthogonal matrix according to Equation 9. To obtain a sparse and structured estimation of $B$, we update each column in it alternatively. Specifically, considering one column of $B$, $B_{:i}$, as an unknown and treating the rest columns of $B$ and the whole of $U$ as constants, we reformulate an optimization problem in Equation 10 to estimate $B_{:i}$:

$$\text{minimize } F(B_{:i}) = ||W^* - \sum_{i \neq j} B^*_{:j} U^{*T}_{:j} - B_{:i} U^{*T}_{:i}||^2_F.$$

$$\text{subject to } B_{ki} = 0 \text{ if } k\text{th score is not associated with } i\text{th latent model.} \qquad (10)$$

Denote $R^* = W^* - \sum_{i \neq j} B^*_{:j} U^{*T}_{:j}$, the solution of $B_{:i}$ in Equation 10 can be obtained by assigning $U^*_{:i} * R^*_{k:}$ to the non-zero elements in $B_{:i}$, where * shows the summation of element-wise multiplication. Each column in the coefficient matrix can be updated alternatively using the same approach.

The problem of estimating $U$ can be formulated as an orthogonal Procrustes problem (Gower, 2004) as the following, we can be efficiently solved by the singular value decomposition:

$$\text{minimize } F(U) = ||W^* - B^* U||^2_F, \text{ subject to } UU^T = I \qquad (11)$$

This problem can be regarded as finding the nearest orthogonal matrix to a given matrix $S = B^{*T} W^*$. The singular value decomposition algorithm can be used to solve it. Algorithm 3 summarizes the steps of estimating $B$, $U$, and $V$ using the proposed BCD-based algorithm.

After estimating $B$ and $U$, the ensemble weights are computed using the algorithm presented in Algorithm 4. The values of $\boldsymbol{b}$ are obtained directly from estimated matrix $B$ and $\sqrt{\frac{1-\pi}{\pi}} \mu^0$s are estimated as the elements of eigenvectors corresponding to the leading eigenvalue of $\Lambda$.

---

**Algorithm 3: Estimating $B$, $U$, and $V$**

**Input:** $R$
1. Initialize $\Theta_0$, $U_0$, and $B_0$.
2. For $j = 1$ to $n\_iteration$:
3.     Estimate $W_j$ as its columns are the first $K$ leading eigenvectors of $R - \Theta_j$ and $V_j$ as a diagonal matrix which its diagonal elements are the first K leading eigenvalues of $R - \Theta_j$.
4.     for $i = 1 \text{ to } K$:
        Compute the nonzero elements of $B^*_{j:i}$ as $B^*_{j_{ki}} = U_{j_{:i}} * R^*_{j_{k:}}$.
    End for
5.     Compute $S = B^T_j W_j$.
6.     Compute singular value decomposition of $S = D\Sigma F^T$.
7.     Update $U_j$ as $U_j = DF^T$.
8.     Update $\Theta_j$ as $diag(\Theta_j) = diag(R - B_j U_j V_j (B_j U_j)^T)$.
9. End for.

**Output:** Optimal $B, U, V$.

---

**Algorithm 4: Estimating model parameters required to compute ensemble weights using SUEL.MF**

**Input:** The assignments ($C_k, k \in \{1, \ldots, K\}$), prediction scores ($f_i(x_t), i \in \{1, \ldots, M\}, t \in \{1, \ldots, N\}$), number of latent groups ($K$), $B, U, V$.

1. Compute the elements of $\boldsymbol{b}$ from $B$.
2. Compute $\sqrt{\frac{1-\pi}{\pi}} \mu_k^0$s as the eigenvectors corresponding to the leading eigenvalue of $UVU^T$.
3. Estimate the accuracy of each predictive score as $w_i = -\frac{b_i \mu_k^0}{\pi - (1-\pi) b_i^2 {\mu_k^0}^2}$.

**Output:** Compute the ensembled scores as $g(x_t) = \sum_{i=1}^{M} w_i f_i(x_t)$.

# 3. Simulation

In order to illustrate the effectiveness of our proposed method, we conducted a simulation study for model performance evaluation and compared the performance of the proposed method with other benchmark models.

## *3.1 Simulation process*

The simulation study encompasses three distinct scenarios, each summarized as follows. In Scenario 1, eight conditionally independent predictors are simulated. The conditional independent predictors can be considered as a special case of the proposed method where the number of latent models equals to the number of predictors and each predictor is assigned to one latent model. Thus, the proposed method still can deal with it, but the advantages of the proposed method will not be reflected in this scenario. To demonstrate the advantages of the proposed method in capturing latent group structure among predictors and distinguishing their discriminability and stability, we further consider four latent models with different discriminability and stability in Scenario 2. The first latent model (C1) is the most accurate one as it has both high discriminability (i.e., difference of component-specific means) and stability (i.e., variance). The second latent model (C2) has low discriminability but high stability. The third latent model (C3) has high discriminability but low stability, and the fourth latent model (C4) is the least accurate one as it has both low discriminability and stability. 9, 6, 7, and 5 predictors are randomly simulated from each latent model with different strength of correlation. Scenario 3 simulates two unbalanced latent models with one latent model (C1) associated with much larger number of predictors (20 predictors) than the other one (2 predictors in C2).

400 samples are simulated in each scenario, whose binary outcomes ($y_t$) are randomly generated using a Bernoulli distribution with the prior probability of belonging to class 1 set at 0.8. Conditioning on the outcome of each sample, its scores from latent models, denoted as $h_k(x_t|y_t)$, are generated randomly using a two-component Gaussian distribution with pre-determined component-specific means and variances as summarized in Table 1. The variances in the two components are equal. The strength of correlation between latent models and each predictor ($b_i$) is generated under constraints to ensure that the weights estimated using Property 2 are identical with their maximal likelihood estimations in Property 1. The random noises ($\varepsilon_i$) are generated randomly using standard normal distribution, i.e. $\varepsilon_i \sim N(0,1)$. Therefore, the predictive score of sample $t$ from $i$th predictor is simulated by $f_i(x_t) = b_i h_k(x_t|y_t) + \varepsilon_i$. Finally, the predictive scores of 400 samples from each predictor are standardized to have mean of 0 and variance of 1.

Table 1. Component-specific means and variances for each scenario.

| Scenario 1 | Scenario 2 | Scenario 3 |
|---|---|---|
| $\mu_k^0$ = [-5, -1, -5.2, -1.04, -5, -1, -5.2, -1.04] $\mu_k^1$ = [1.25, 0.25, 1.3, 0.26, 1.25, 0.25, 1.3, 0.26] $\sigma_k$ =[4, 5, 6, 7, 8, 9, 10, 11] | $\mu_k^0$ = [-5, -1, -5.2, -1.04] $\mu_k^1$ = [1.25, 0.25, 1.3, 0.26] $\sigma_k$ =[6, 6, 3, 3] | $\mu_k^0$ = [-5, -1] $\mu_k^1$ = [1.25, 0.25] $\sigma_k$ =[5, 2] |

## 3.2 Benchmark models and evaluation metrics

The prediction performance of SUEL.CQO and SUEL.MF approaches is compared with three benchmark models, Eigen model, the average-based unsupervised ensemble learning, and the ground truth. The Eigen model assumes that each predictor has a two-component Gaussian distribution and its ensemble weight is proportional solely to the difference of component-specific means. It neglects the influence of component-specific variance and considers only the covariance among pairs of predictors belonging to different blocks (Ionita-Laza et al., 2016). Notably, the proposed SUEL.MF and SUEL.CQO approaches explicitly leverage predictor dependencies and enhance accuracy estimation by considering both distinguishability and stability. Therefore, they hold promise for outperforming the Eigen method, which does not account for these factors in the same manner. In addition, an average-based unsupervised ensemble learning model is also compared, which simply takes the average of predictive scores. Furthermore, a ground truth model, which uses the ground truth accuracy of each predictor as the ensemble weight is compared. Lastly, the best individual predictor's performance is incorporated for model comparison.

The prediction accuracy of each ensemble model is evaluated through area under ROC (ROC-AUC), area under Precision-Recall curve (PRC-AUC) and the decile analysis. In the decile analysis, the samples are segmented into 10 deciles based on the rank of their predicted ensemble scores. The distribution of positive samples across 10 deciles is visualized using the histogram. More positive samples allocating in the first decile indicates a more accurate prediction of the ensemble scores. To test the goodness-of-fit of the proposed method, the estimated weights are compared with the ground truth weights in each scenario.

*3.3 Results*

The optimal number of latent models and the assignment of predictors to different latent models are identified by the elbow method using the principal component analysis of sample covariance matrix and the hierarchical clustering on the quartet-based similarity matrix, respectively. The number of latent models and the assignment of scores are consistent with our pre-identified structures. Figure 3 shows the percentage of variance explained by different number of principal components in all scenarios. The knee of the curve represents the optimal number of latent variables which is equal to 8, 4, and 2 for Scenario 1, Scenario 2, and Scenario 3, respectively. Then, the number of latent models is set to the optimal number of latent models and optimal assignment of predictors to latent models are obtained using hierarchical clustering on the quartet-based similarity matrix. Figure 4 visualizes the structure of the quartet-based similarity matrix in all scenarios, clearly showing higher similarities among scores within each latent model and lower similarities between scores from different latent models.

The prediction performance of our proposed models (SUEL.MF and SUEL.CQO) and four baseline models under different scenarios is summarized in Table 2. In all scenarios, SUEL.MF outperforms the other ensemble approaches while SUEL.CQO is the second-best approach. The performance of the proposed approaches is also close to the ground truth model, which further demonstrates the effectiveness of the proposed algorithm. In addition, the estimated weights using SUEL.MF approach showcases high correlations with the true weights in all scenarios (Figure 5). Furthermore, in Scenario 2, we observe that predictors associated with the latent model C1, which has both high discriminability and stability, are ranked highest and the scores associated with the latent model C4, which has both low discriminability and stability, are ranked lowest in the proposed method. This observation is consistent with our prior knowledge of the simulation model. Figure 6-8 show the distribution of positive samples across different decile of prediction scores. In all scenarios, the proposed methods (MF and SUEL-CQO) have higher number of positive samples in the first few deciles of prediction scores compared to the benchmark models, which demonstrates that the scores predicted from the proposed methods can better distinguish the positive samples with negative ones. Moreover, the MF-based algorithm can achieve the same accuracy with the ground truth in the first decile.

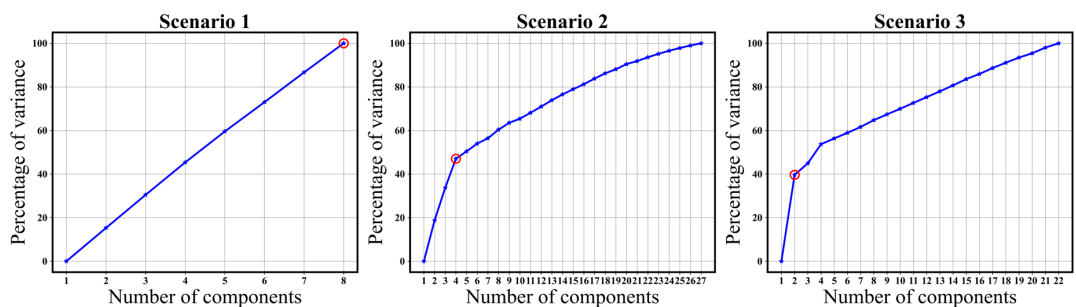

Figure 3. The percentage of explained variance versus the number of components.

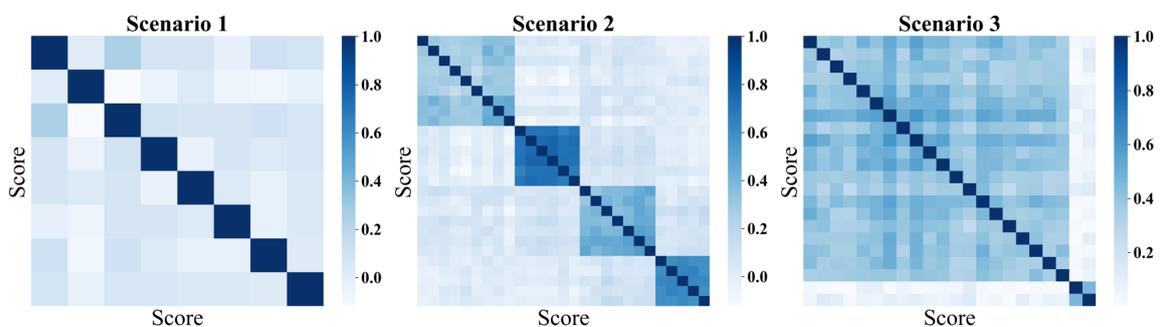

Figure 4. Heatmap of quartet-based similarity matrix

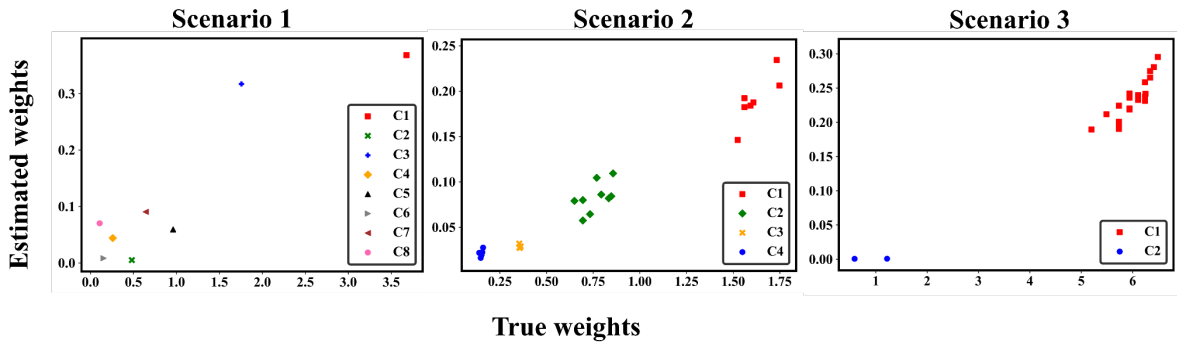

Figure 5. Estimated weights versus true weights in SUEL.MF under different scenarios.

Table 2. The prediction performance of different approaches under three scenarios.

| Prediction models | Scenario 1 | | Scenario 2 | | Scenario 3 | |
| --- | --- | --- | --- | --- | --- | --- |
| | ROC | PRC | ROC | PRC | ROC | PRC |
| SUEL.MF | **0.92** | **0.79** | **0.93** | **0.85** | **0.80** | **0.57** |
| SUEL.CQO | 0.92 | 0.78 | **0.93** | 0.84 | **0.80** | 0.56 |
| Eigen | 0.89 | 0.72 | 0.90 | 0.80 | 0.73 | 0.49 |
| Average | 0.85 | 0.66 | 0.84 | 0.63 | 0.76 | 0.54 |
| Best score | 0.88 | 0.70 | 0.86 | 0.67 | 0.77 | 0.37 |
| *Ground truth* | *0.93* | *0.83* | *0.93* | *0.85* | *0.80* | *0.58* |

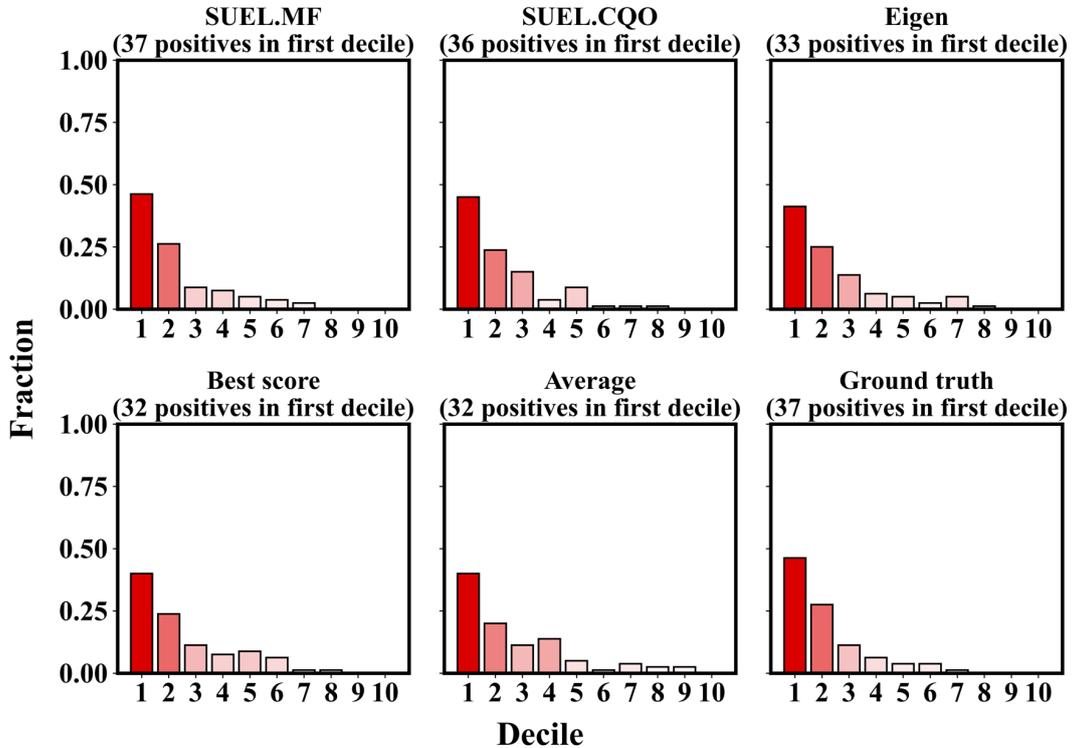

Figure 6. The distribution of positive samples across different decile of prediction models in Scenario 1.

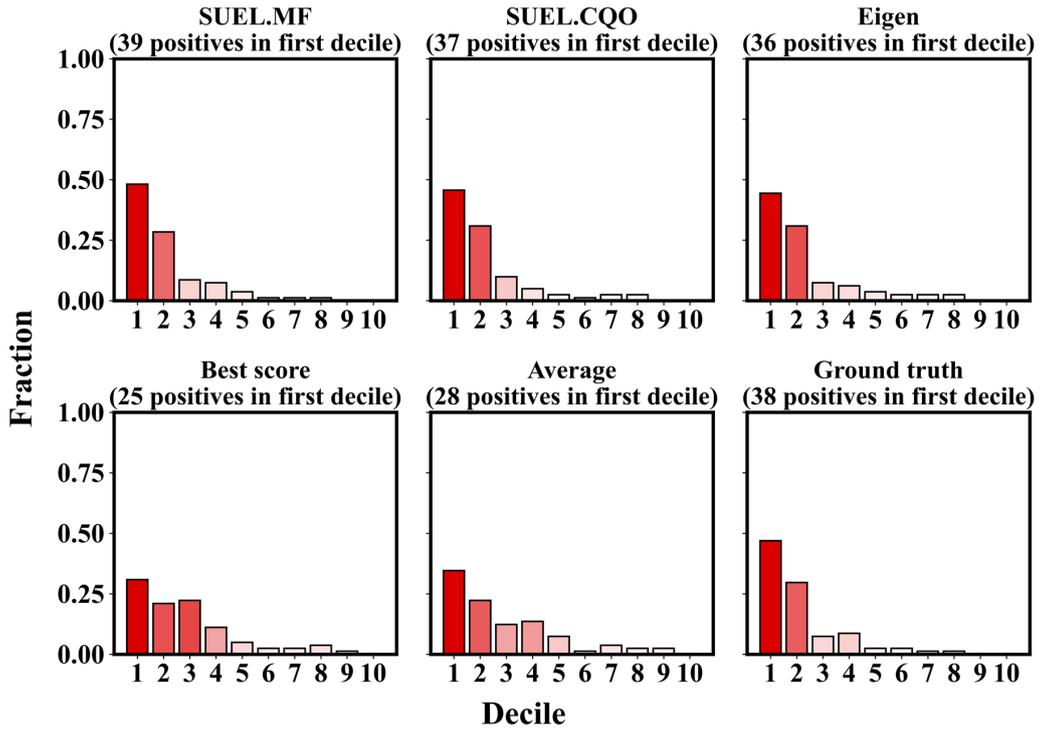

Figure 7. The distribution of positive samples across different decile of prediction models in Scenario 2.

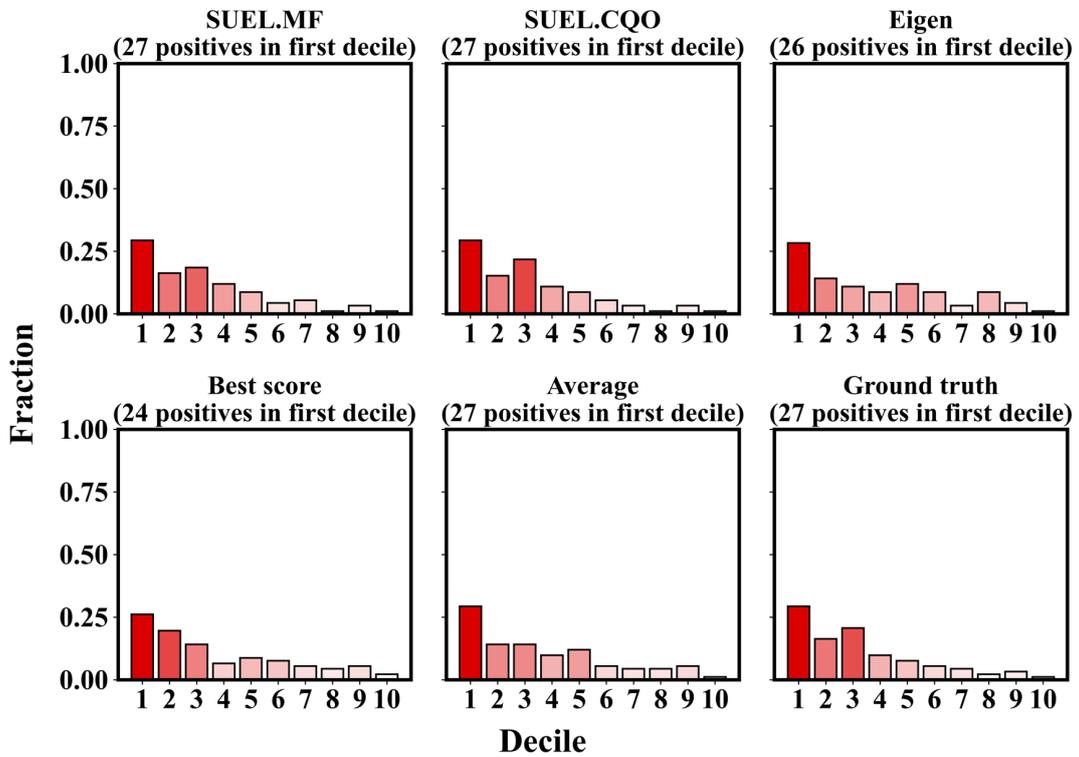

Figure 8. The distribution of positive samples across different decile of prediction models in Scenario 3.

# 4. Case study

## *4.1 Disease-associated risk genes discovery*

Autism Spectrum Disorder (ASD) is a complex neurodevelopmental disorder characterized by the deficits in social communication, impaired language development, and the presence of highly restricted interests and/or stereotyped repetitive behaviors (State & Levitt, 2011). The early diagnosis and treatment of ASD is still a challenge due to the very limited understanding of its underlying pathology. Recent advances in the Genome-Wide Association Studies (GWAS) of large-scale population have demonstrated the role of genetics and genomics in ASD development which are promising to facilitate the development of novel screening approaches and effective personalized medication for ASD (C Yuen et al., 2017; Hallmayer et al., 2011; Turner et al., 2017). However, the knowledge of risk genes associated with ASD is still very far from complete (Lin et al. 2020). Therefore, numerous computational models have been developed in the literature to rank and score the genes by their likelihood of association with ASD (C. Zhang & Shen, 2017); (Iossifov et al., 2015); (Krishnan et al., 2016); (Sanders et al., 2015).

These scoring systems aim to predict the likelihood that each gene is linked to ASD. They incorporate different types of scores, such as statistical scores derived from extensive Genome-Wide Association Studies (GWAS), network-based scores that consider gene interactions, and scores generated using machine learning techniques. Despite these efforts, the performance of existing scoring systems remains below the desired level. Given that various predictive scores can offer complementary insights into disease-gene associations, the prospect of combining these existing predictive scores into a new ensemble score holds significant potential for enhancing the discovery of risk genes associated with ASD. This approach aims to create an ensemble score with improved robustness and accuracy, drawing upon the strengths of multiple predictive measures.

*4.2 Data Description*

In this study, we aggregate 20 ranking scores developed for ASD risk gene discovery in the existing literature which were built under various approaches. These scores encompass 4 machine learning model-based gene ranking scores for ASD, including Zhang D (C. Zhang & Shen, 2017), IOSSIFOV (Iossifov et al., 2015), Krishnan (Krishnan et al., 2016), Sander (Sanders et al., 2015). We also consider 9 gene-level constraint metrics and general gene features developed from the exome data of more than 60,000 individuals from the ExAC to quantify the sensitivity of genes to variations (e.g. Z scores for synonymous which shows whether the gene is more intolerant of variation). As risk genes are usually functionally correlated via gene networks, we further consider 7 gene network's topological scores (e.g. the hubness of each gene in the network) obtained from prior research to rank the genes based on their functional connectivity (Lin, Afshar, Rajadhyaksha, Potash, & Han, 2020). The 20 ranking scores are curated for 14,642 genes in total. Among them 1,016 genes were previously implicated to be associated/independent with the ASD (115 are associated with ASD and 901 are independent with ASD). We use all 14,642 genes to train the proposed unsupervised ensemble learning model and 1,016 labeled genes to evaluate the model performance. Additionally, we conduct an external evaluation using a set of 214 independent ASD candidate genes. All scores are standardized with a mean of zero and a standard deviation of one.

*4.3 Evaluation metrics*

To evaluate the effectiveness of our proposed method, we employ various metrics including ROC-AUC and PRC-AUC, along with a downstream analysis. The downstream analysis involves computing the enrichment of ASD risk genes within each decile of the sorted predicted scores from different models. We anticipate that more accurate predictions will yield a higher fraction of ASD genes in the first decile. In our evaluation, we compare the prediction performance of the proposed approaches (SUEL.MF and SUEL.CQO) to several other models: Eigen, the average-based ensemble score, and the best prediction performance obtained from individual scores. Furthermore, we conduct a sensitivity analysis to gauge how inaccurately estimating the number of latent models can impact the prediction performance of our proposed models and the benchmark models. Specifically, we compare the prediction performance of SUEL.MF and Eigen models by varying the assignment of predictors to different numbers of latent models.

*4.4 Results*

To identify the optimal number of latent models, we employ the elbow method and principal component analysis on the covariance matrix of standardized scores. As illustrated in Figure 9, we determine that three latent models provide the best fit for our data. Subsequently, leveraging the quartet-based similarity matrix, showcased in Figure 10, we employed hierarchical clustering to categorize the 20 ranking scores into three latent models. It becomes evident that the first group predominantly consists of ASD ranking scores found in existing literature, the second group is comprised of gene-level constraint metrics, and the third group exclusively encompasses topological features.

With the latent model structure identified, we proceeded to estimate model parameters using our two proposed algorithms. Our evaluation involved comparing the prediction performance of SUEL.CQO and SUEL.MF with three benchmark models, utilizing the test gene dataset. The results in Table 3 demonstrate that, SUEL.MF outperforms the benchmark models with respect to the ROC-AUC. However, the Sander score has higher PRC-AUC compared to the proposed method. It's worth mentioning that the Sander score was trained using a dataset that highly overlaps with our test data, encompassing around 50% of the test genes, possibly introducing bias. Figure 11 illustrates the estimated weights assigning to each ranking system. ExaC, IOSSIFOV and IOF_Z scores have the highest estimated weights among different ranking scores in the model. Thus, these ranking systems showcase the highest accuracy compared to other ranking scores in the ASD gene prediction. This finding can help us to assess the prediction performance of different ranking scores without using any labeled data.

In Figure 12, our downstream analysis evaluates predictions from different methods using external independent data. The results indicate that ASD-associated genes were most enriched in the first decile of scores predicted by the proposed SUEL.MF method, demonstrating its superior performance in ranking ASD-associated risk genes compared to the other three models.

Furthermore, our sensitivity analysis reveals that while the SUEL.MF algorithm outperforms the Eigen method under the optimal number of latent models (three latent models), the prediction performance of SUEL.MF significantly declines in terms of ROC-AUC and PRC-AUC when latent models are inaccurately determined (see Figure 13). This is in contrast to the Eigen approach, which exhibited lower sensitivity to the number of latent models, as it does not exploit the latent structure in model estimation by assuming the scores are conditional independent. In the SUEL.MF approach, however, changing the number of latent models leads to the changes in model structure and number of parameters, which further leads to the significant drop in prediction performance of the model. Therefore, it is critical to find the optimal latent structure in the correlated scores using the proposed algorithm.

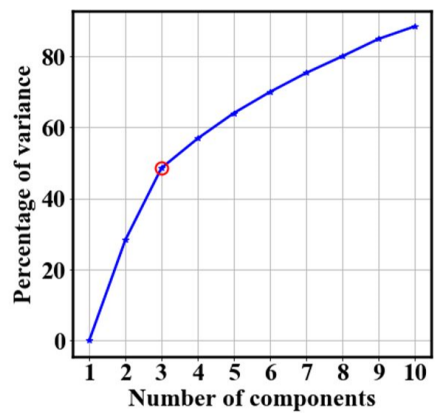

Figure 9. The percentage of explained variance versus number of principal components.

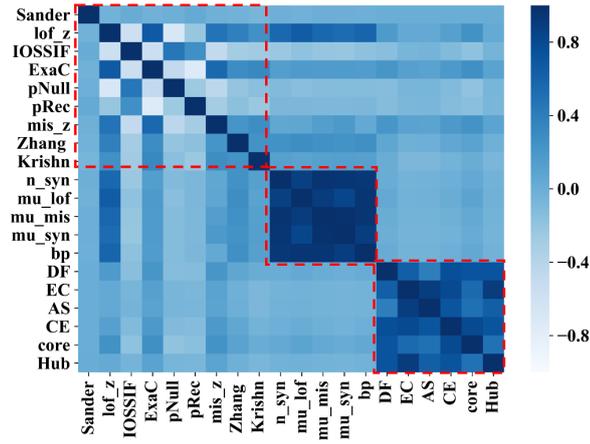

Figure 10. The heatmap of quartet-based similarity matrix for ASD case study.

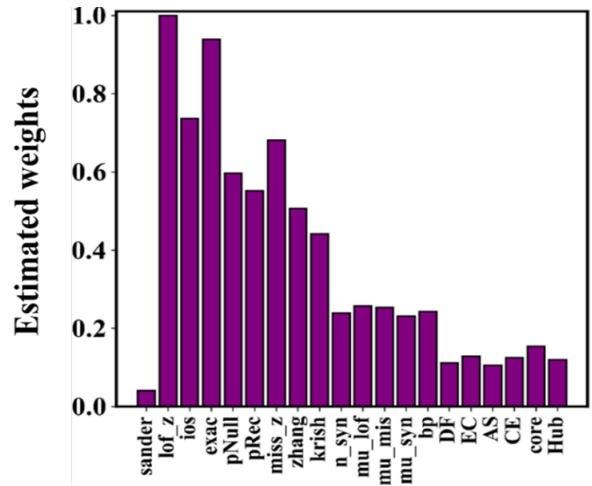

Figure 11. The estimated weights of SUEL.MF approach for ranking systems.

Table 3. The prediction performance of different approaches.

|         | SUEL.CQO | Eigen | SUEL.MF | Average | Best score |
|---------|----------|-------|---------|---------|------------|
| ROC-AUC | 0.84     | 0.82  | **0.84** | 0.66   | 0.78       |
| PRC-AUC | 0.42     | 0.42  | 0.47    | 0.20    | **0.67**   |

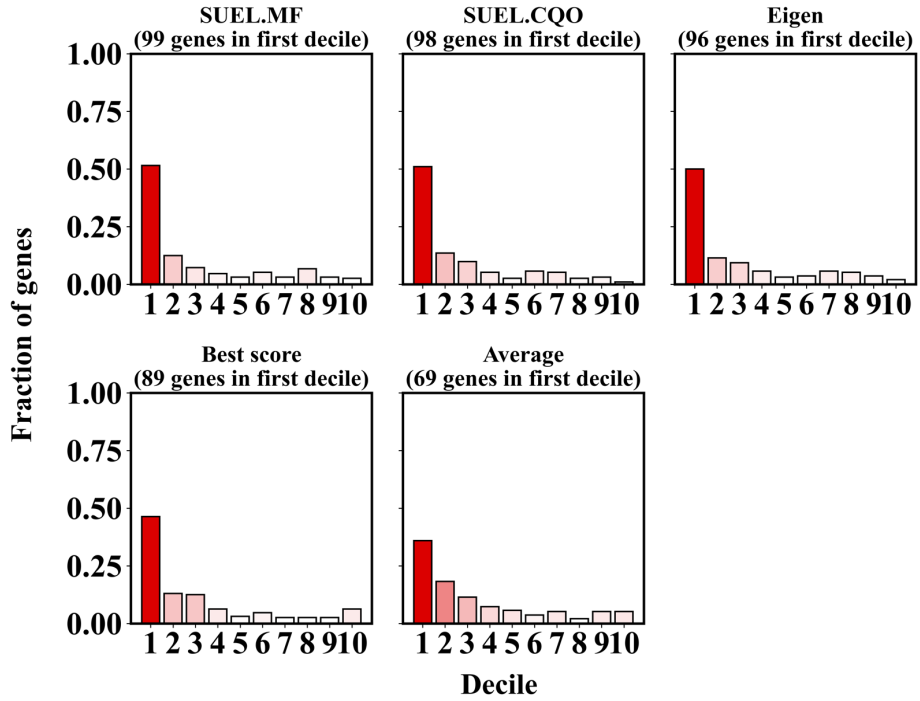

Figure 12. The distribution of ASD genes across different decile of prediction models.

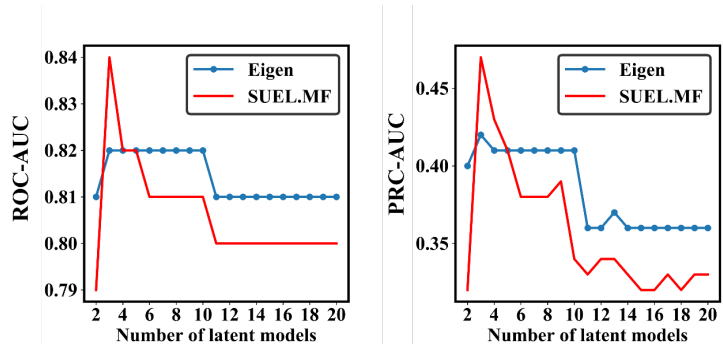

Figure 13. The impact of number of latent models in the accuracy of different models.

## 5. Conclusion

In conclusion, we have developed a novel Structured Unsupervised Ensemble Learning (SUEL) approach designed to seamlessly integrate dependent and continuous predictive scores in absence of labeled data. The proposed method estimates the unknown accuracy of each predictor by considering the latent structure among predictors, the distinguishability (mean difference) and the stability (variance) of each predictor. Two novel correlation-based decomposition algorithms, SUEL.CQO and SUEL.MF, have been proposed to find the optimal latent structure and estimate the model parameters. To evaluate the proposed method, the prediction performance of SUEL.MF and SUEL.CQO are compared to other benchmark models using both simulated data and real-world data of ASD risk gene prediction. The results obtained from both simulated and real-world data demonstrate that SUEL.MF outperformed other benchmark models. Moreover, this approach can efficiently discover the dependency between predictors, accurately estimate the accuracy of each predictor, and optimally integrate the predictors to an ensemble model which leads to the most significant improvement in the prediction accuracy compared to the individual predictors.

There are some limitations of this work. First, the proposed method can be applied to various unsupervised ensemble learning problems in the engineering systems. In the future, we will study and evaluate its effectiveness in the other domains, such as ranking and integrating various in-situ sensors for fault prognostics in the manufacturing system (Kolokas, Vafeiadis, Ioannidis, & Tzovaras, 2020) and ranking and integrating multiple clustering structures for consensus clustering (Ünlü & Xanthopoulos, 2019). Second, the proposed method is developed to rank and combine the scores from two balanced classes. In the future, we will extend the proposed model to predict more than two classes. Imbalanced classification problem is commonly observed in real applications, such as the abnormal events are much less than the normal ones in engineering systems. To solve these imbalanced classification problem, we will leverage the extreme event distributions to capture the latent models and their structure in the proposed method. Last but not least, the proposed method can be further extended to rank and combine scores coming from more complex latent structures, such as the tree structures.

## Acknowledgements

This research was funded by NIMH, R01MH121394

Qi, Y. (2012). Random forest for bioinformatics. In *Ensemble Machine Learning* (pp. 307-323): Springer.

Sanders, S. J., He, X., Willsey, A. J., Ercan-Sencicek, A. G., Samocha, K. E., Cicek, A. E., . . . Dong, S. (2015). Insights into autism spectrum disorder genomic architecture and biology from 71 risk loci. *Neuron, 87*(6), 1215-1233.

Schapire, R. E. (2013). Explaining adaboost. In *Empirical Inference* (pp. 37-52): Springer.

Shaham, U., Cheng, X., Dror, O., Jaffe, A., Nadler, B., Chang, J., & Kluger, Y. (2016). *A deep learning approach to unsupervised ensemble learning.* Paper presented at the International conference on machine learning.

Song, C., & Liu, K. (2018). Statistical degradation modeling and prognostics of multiple sensor signals via data fusion: A composite health index approach. *IISE Transactions, 50*(10), 853-867.

State, M. W., & Levitt, P. (2011). The conundrums of understanding genetic risks for autism spectrum disorders. *Nature Neuroscience, 14*(12), 1499-1506.

Tama, B. A., & Lim, S. (2021). Ensemble learning for intrusion detection systems: A systematic mapping study and cross-benchmark evaluation. *Computer Science Review, 39*, 100357.

Traganitis, P. A., Pages-Zamora, A., & Giannakis, G. B. (2018). Blind multiclass ensemble classification. *IEEE Transactions on Signal Processing, 66*(18), 4737-4752.

Tsogbaatar, E., Bhuyan, M. H., Taenaka, Y., Fall, D., Gonchigsumlaa, K., Elmroth, E., & Kadobayashi, Y. (2020). *SDN-enabled IoT anomaly detection using ensemble learning.* Paper presented at the Artificial Intelligence Applications and Innovations: 16th IFIP WG 12.5 International Conference, AIAI 2020, Neos Marmaras, Greece, June 5–7, 2020, Proceedings, Part II 16.

Turner, T. N., Coe, B. P., Dickel, D. E., Hoekzema, K., Nelson, B. J., Zody, M. C., . . . Pennacchio, L. A. (2017). Genomic patterns of de novo mutation in simplex autism. *Cell, 171*(3), 710-722. e712.

Ünlü, R., & Xanthopoulos, P. (2019). A weighted framework for unsupervised ensemble learning based on internal quality measures. *Annals of Operations Research, 276*, 229-247.
36